\documentclass[conference]{IEEEtran}
\IEEEoverridecommandlockouts
\usepackage{cite}
\usepackage{amsmath,amssymb,amsfonts}
\usepackage{algorithmic}
\usepackage{graphicx}
\usepackage{textcomp}
\usepackage{xcolor}
\usepackage{booktabs}
\usepackage{wrapfig}
\usepackage{times}
\usepackage{soul}
\usepackage{url}
\usepackage[hidelinks]{hyperref}
\usepackage[utf8]{inputenc}
\usepackage[small]{caption}
\usepackage{graphicx}
\usepackage{amsmath}
\usepackage{amsthm}
\usepackage{booktabs}
\usepackage{algorithm}
\usepackage{algorithmic}
\usepackage[switch]{lineno}
\usepackage{color}
\usepackage{multirow}
\usepackage{bbding}
\usepackage{pifont}
\usepackage{wasysym}
\usepackage{amssymb}
\usepackage{amsfonts,amssymb}
\usepackage{colortbl}

\definecolor{mygray}{gray}{.9}

\usepackage{titlesec}

\usepackage[capitalize]{cleveref}
\crefname{section}{Sec.}{Secs.}
\Crefname{section}{Section}{Sections}
\Crefname{table}{Table}{Tables}
\crefname{table}{Tab.}{Tabs.}
\crefname{algocf}{alg.}{algs.}

\def\BibTeX{{\rm B\kern-.05em{\sc i\kern-.025em b}\kern-.08em
    T\kern-.1667em\lower.7ex\hbox{E}\kern-.125emX}}
\begin{document}

\title{
SUEDE: \textbf{S}hared \textbf{U}nified \textbf{E}xperts for Physical-\\Digital Face Attack \textbf{D}etection \textbf{E}nhancement}

\author{
\IEEEauthorblockN{
Zuying Xie$^{1,}$\IEEEauthorrefmark{1},
Changtao Miao$^{2,}$\IEEEauthorrefmark{1},
Ajian Liu$^{3,}$\IEEEauthorrefmark{2},
Jiabao Guo$^{1}$, 
Feng Li$^{1}$, 
Dan Guo$^{1}$, 
and Yunfeng Diao$^{1,}$\IEEEauthorrefmark{2}
}

\IEEEauthorblockA{$^{1}$Hefei University of Technology, Hefei, China}
\IEEEauthorblockA{$^{2}$University of Science and Technology of China, Hefei, China}
\IEEEauthorblockA{$^{3}$Institute of Automation Chinese Academy of Sciences, Beijing, China}

\small{
2024110543@mail.hfut.edu.cn,miaoct@mail.ustc.edu.cn,
\{garbo\_guo,fengli,guodan\}@hfut.edu.cn,
ajian.liu92@gmail.com, diaoyunfeng@hfut.edu.cn
}





\thanks{\IEEEauthorrefmark{1}Equal contribution.
        \IEEEauthorrefmark{2}Corresponding author.}
}
\maketitle

\begin{abstract}

Face recognition systems are vulnerable to physical attacks (e.g., printed photos) and digital threats (e.g., DeepFake), which are currently being studied as independent visual tasks, such as Face Anti-Spoofing and Forgery Detection. The inherent differences among various attack types present significant challenges in identifying a common feature space, making it difficult to develop a unified framework for detecting data from both attack modalities simultaneously. Inspired by the efficacy of Mixture-of-Experts (MoE) in learning across diverse domains, we explore utilizing multiple experts to learn the distinct features of various attack types. However, the feature distributions of physical and digital attacks overlap and differ. This suggests that relying solely on distinct experts to learn the unique features of each attack type may overlook shared knowledge between them.  
To address these issues, we propose \textit{SUEDE}, the \textbf{S}hared \textbf{U}nified \textbf{E}xperts for Physical-Digital Face Attack \textbf{D}etection \textbf{E}nhancement. SUEDE combines a shared expert (always activated) to capture common features for both attack types and multiple routed experts (selectively activated) for specific attack types. Further, we integrate CLIP as the base network to ensure the shared expert benefits from prior visual knowledge and align visual-text representations in a unified space. Extensive results demonstrate SUEDE achieves superior performance compared to state-of-the-art unified detection methods.
\end{abstract}


\section{Introduction}
\label{sec:intro}
As one of the most successful computer vision technologies, face recognition has been widely applied in various fields such as face payment and video surveillance. However, the robustness of these systems has been questioned recently, which shows that they are susceptible to both physical attacks\cite{liu2021casia,liu2022contrastive} and digital attacks\cite{cdf}. The former involves presenting a face on a physical medium in front of the camera to deceive the face recognition system, while the latter employs imperceptible visual manipulation in the digital domain to fool the face recognition system.

\begin{figure}[t]
\centering
\resizebox{1.0\linewidth}{!}{
\includegraphics[width=1\linewidth]{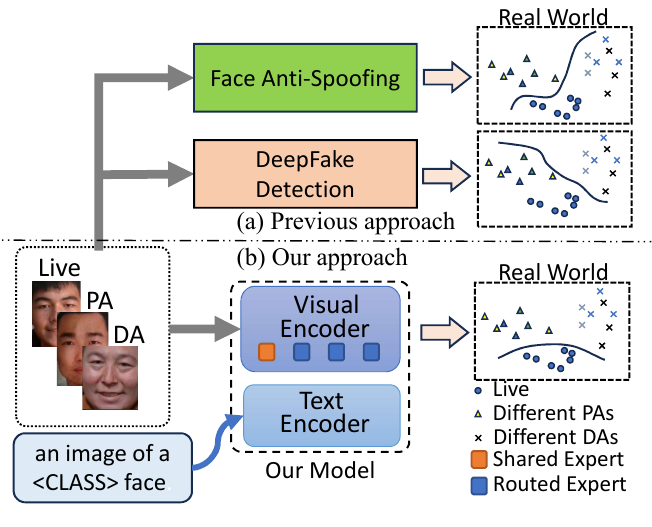}
}
\caption{A high-level illustration of our proposed method. The existing work designs different detection methods for physical attacks (PAs) and digital attacks (DAs) separately, failing in real-world unified face attacks. In contrast, our method can detect both PAs and DAs simultaneously via shared unified experts.}
\label{fig:high_level}
\vspace{-5mm}
\end{figure}

Physical attacks (PAs) have always existed, including video replay attacks \cite{2d} and 3D mask attacks \cite{liu2022contrastive}, among others. Face anti-spoofing (FAS) is a technique designed to detect such physical attacks. 
Today, FAS can leverage multiple modalities (e.g., depth, rPPG, RGB images) and design deep learning networks that autonomously learn to detect real and fake faces\cite{ijcai2022p165,Liu2023FMViTFM,liu2021face}. Digital attacks (DAs) have become increasingly powerful with the development of generative models, enabling the easy generation or tampering of faces using techniques such as deepfake \cite{ciftci2021detection}, SimSwap~\cite{simswap}, etc. Forgery detection refers to the process of identifying digital attacks and it can be broadly categorized into spatial detectors and frequency detectors: spatial detectors focus on specific representations, such as the location of forged regions \cite{fd_forgeryArea, mct_location}, disentangled learning \cite{fd_disentagled}, and image reconstruction\cite{fd_imgRecon}, while frequency detectors address this issue by focusing on the frequency domain for forgery detection \cite{fd_f1, mct_frequency_1, mct_frequency_2}.

Both physical and digital attacks pose substantial threats to the security of face recognition systems. Due to the distinct characteristics of these attacks, existing studies typically treat them as separate issues \cite{liu2024moeit,ffd}, as shown in \cref{fig:high_level}. However, real-world attacks often involve a combination of both, rendering single detection methods ineffective in real-world scenarios~\cite{fang2024unified,liu2024cfpl,liu2024moeit}. Furthermore, treating these attacks independently necessitates deploying multiple models, leading to increased computational costs, which also fails to provide a unified framework for detecting fake faces from both attack modalities simultaneously. Consequently, developing a unified attack detection (UAD) framework is necessary and urgent. Very recently, researchers have attempted to integrate datasets of physical attacks and digital attacks, and design a unified framework to enhance the performance of UAD\cite{yu2024benchmarking, fang2024unified, zou2024softmoe}. 
These studies try to map both physical and digital attacks to the same unified feature space. However, due to the inherent differences between physical and digital attacks, it is still challenging to accurately distinguish real faces while categorizing these two types of attacks within the same ``fake" latent space. 

Considering that Mixture-of-Experts (MoE)\cite{jacobs1991moe1,dai2024deepseekmoe} 
facilitates the acquisition of attack-related knowledge by enabling experts to learn the diversity of attacks, we investigate its potential for addressing the unified attack detection (UAD) task. However, our preliminary experiments(as reported in \cref{fig:vit+moe}) found that directly applying MoE to the UAD task does not result in significant performance improvement. First, as illustrated in \cref{fig:tsne}, the latent feature distribution of physical attacks and digital attacks includes both unrelated and overlapping regions. This implies that only utilizing distinct experts to learn unique features of specific attack types may ignore the shared knowledge between both attack types. Moreover, MoE was originally designed for natural language processing (NLP)\cite{shazeer2017outrageously}, where sparse and sequential inputs are more amenable to expert specialization. In contrast, visual face data often contains rich spatial information that MoE struggles to exploit without architectural modifications or additional mechanisms. To address these limitations, we propose \textit{SUEDE}, the \textbf{S}hared \textbf{U}nified \textbf{E}xperts for Physical-Digital Face Attack \textbf{D}etection \textbf{E}nhancement. SUEDE consists of a shared expert (always activated) and multiple routed experts (selectively activated). The shared expert captures common features across various attack types, while the routed experts focus on learning the unique features of specific attack types. To fully unleash the power of Shared Unified Experts in visual UAD tasks, we employ CLIP as the base network to ensure that the shared unified experts inherit prior visual knowledge and align visual-text representations within a shared embedding space. Extensive results demonstrate that SUEDE effectively enhances unified detection performance, particularly in cross-domain attacks, thereby demonstrating its flexibility and effectiveness in addressing unified attack detection.

\section{Preliminaries}
\subsection{Mixture-of-Experts (MoE)}
The MoE  \cite{shazeer2017outrageously} architecture typically comprises multiple experts and a gating network. Experts are distinct neural network layers with varying parameters and specific functions, such as multi-layer perceptrons (MLPs)\cite{dai2024deepseekmoe}. In different application scenarios, the structure and parameters of these experts can be tailored to the task at hand. The gating network determines which experts to activate and how to combine their outputs based on the input data. This network can be implemented as a simple linear layer followed by a softmax activation function, which converts the gating network output into a probability distribution. This ensures that each element of the weight vector $ w $ is within the interval $[0, 1]$ and that the sum of the weights equals 1.

\subsection{Text-Image Pre-trained CLIP Model}
CLIP\cite{radford2021CLIP} comprises an Image Encoder and a Text Encoder. The Image Encoder, utilizing architectures such as ResNet or Vision Transformer, transforms input images into meaningful feature vectors. The Text Encoder, typically based on the Transformer architecture, tokenizes text and processes it to generate a semantic feature vector, employing multi-head attention to capture contextual relationships.
Central to CLIP is the contrastive learning objective. During training, the encoders produce feature vectors for image-text pairs, from which similarity scores are computed. The objective is to maximize the similarity of correct pairs while minimizing that of incorrect ones, thereby learning a semantic alignment between images and texts.
The loss function for CLIP is based on the cross-entropy of text-image similarity, as described in:
\begin{equation}
 L_{CE} = -\frac{1}{N} \sum_{i=1}^{N} \log \frac{\exp(S_{ii})}{\sum_{j=1}^{N} \exp(S_{ij})} .\quad \label{clip}
\end{equation}

Here, $ S $ denotes the similarity matrix, where $ S_{ij} $ represents the similarity between the embedding of the $ i^{th} $ image and the $ j^{th} $ text embedding, and $ N $ is the number of sample pairs.

\section{Method}

\subsection{Motivation}

\begin{figure}[!ht]
\centering
\includegraphics[width=1\linewidth]{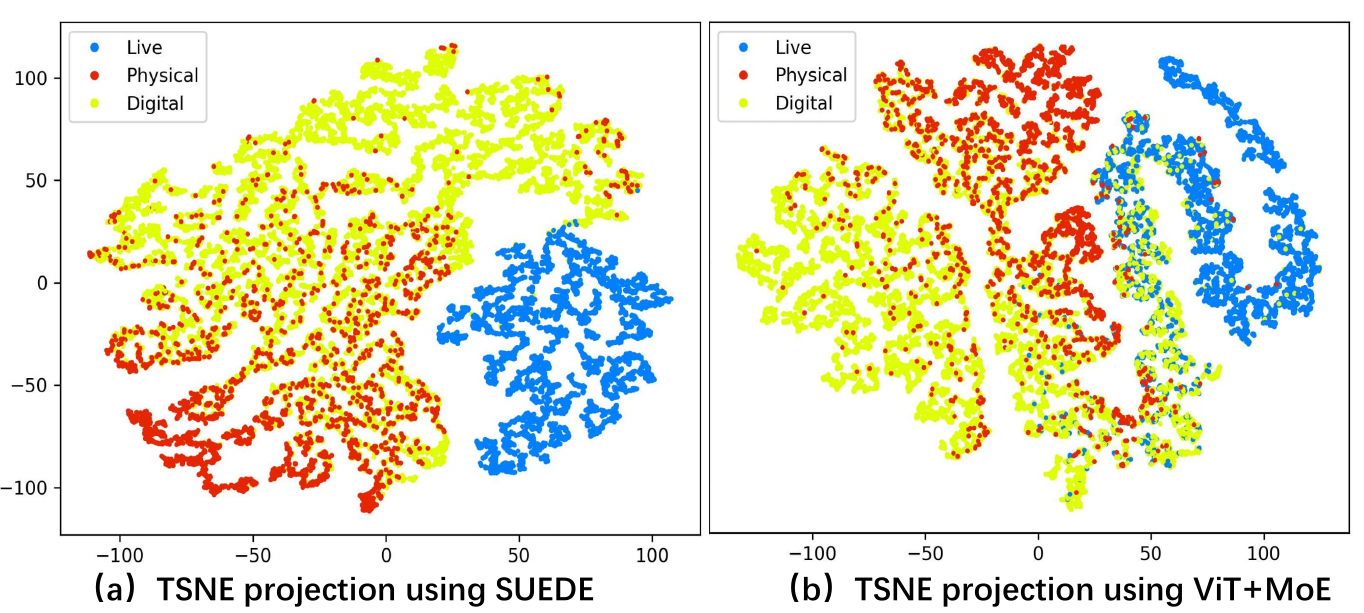}
\caption{The feature distribution of the unified attack protocol in the UniAttackData using SUEDE(a) and ViT+MoE(b), respectively. Live means real face data. ``Physical" and ``Digital" refer to the fake data generated by physical attacks and digital attacks, respectively.}
\label{fig:tsne}
\end{figure}
The goal of the UAD task is to develop a unified detection framework for both physical and digital face attacks. However, the inherent differences between physical and digital attacks, combined with the variety of attack methods within each category, make it challenging to identify fake faces generated by diverse attack types. MoE mechanism allows multiple experts to make distinct judgments through adaptive activation based on input data rather than relying on fixed model parameters as in traditional models. This capability makes MoE particularly suitable for handling diverse attack types. To this end, we explore employing MoE to capture the unique features of various attack types. MoE is typically integrated into Transformers by replacing their original FFN layers, making it a natural choice to incorporate MoE into Vision Transformers (ViT) to defend against unified attacks.

\begin{figure}[t]
\vspace{-2mm}
\centering
\scalebox{0.8}{\includegraphics[width=0.6\linewidth]{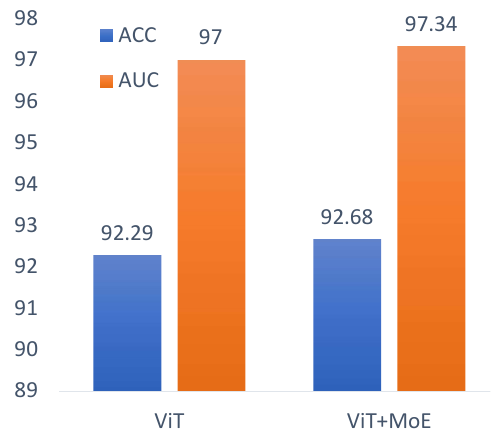}}
\caption{\footnotesize  Comparative experiment between ViT and ViT with MoE. The experiment is conducted on Protocol1 of the UniAttackData dataset.}
\label{fig:vit+moe}
\vspace{-5mm}
\end{figure}

However, our preliminary experiments, as shown in \cref{fig:vit+moe}, indicate that a naive application of MoE for UAD does not have significant performance improvement. To investigate this reason, we utilize TSNE~\cite{tsne} to visualize the feature distribution of physical face attacks, digital face attacks and real faces in \cref{fig:tsne} (b). The latent feature distribution reveals that physical and digital attacks include both unrelated and overlapping regions. Distinct features of various attack images, such as depth information for physical attacks and forgery artifacts for digital attacks, can be effectively captured by different experts in a traditional MoE structure. However, the structural features of faces and background information across live, physical, and digital attacks exhibit considerable similarity. This observation indicates that relying solely on distinct experts to learn unique features of specific attack types may overlook the shared knowledge between the two types of attacks. Therefore, we argue that the MoE includes a shared expert capable of learning unified features across both attack types. In addition, MoE was originally designed for NLP\cite{shazeer2017outrageously, dai2024deepseekmoe}. As demonstrated in \cref{fig:vit+moe}, its direct adaption on ViT does not perform well. Thus, we argue that relying solely on a visual encoder as the backbone is insufficient. Instead, aligning visual and text representations is crucial to fully exploit the potential of MoE for the UAD task.

\subsection{Shared Unified Experts (SUE)}

\begin{figure}[!t]
\centering
\includegraphics[width=1\linewidth]{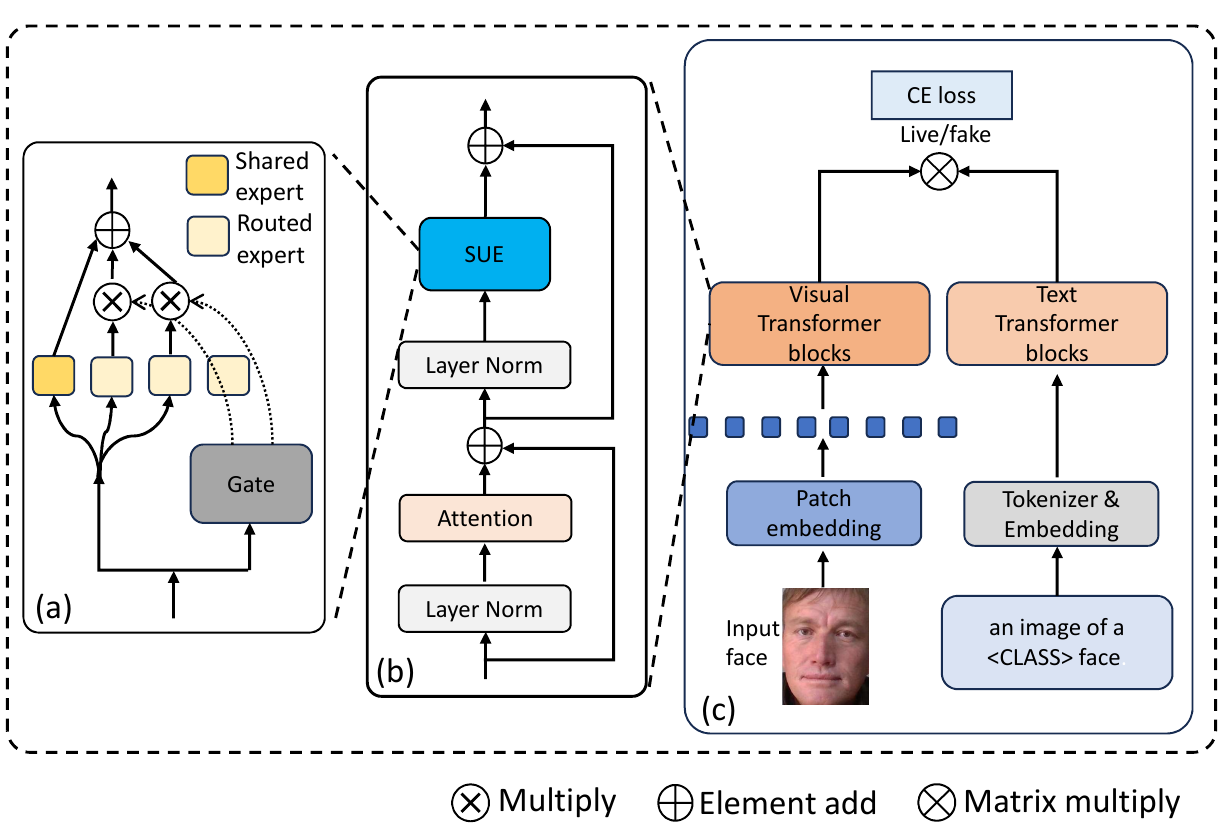}
\caption{The framework of SUEDU. (a) is the design of the Shared Unified Expert. (b) provides a detailed representation of the Transformer block. (c) illustrates the structure of CLIP. 
}
\label{fig:algo}
\vspace{-4mm}
\end{figure}

To address these issues, we propose \textit{SUEDE}, the \textbf{S}hared \textbf{U}nified \textbf{E}xperts for Physical-Digital Face Attack \textbf{D}etection \textbf{E}nhancement. The framework of SUEDE is presented in \cref{fig:algo}. SUEDE consists of two branches: the image branch, which includes an image embedding layer and an image transformer, and the text branch, which comprises a tokenizer, embedding, and a text transformer. In the image branch, we introduce an innovative Shared Unified Expert module that learns common features across various attack images using shared experts, while specialized experts capture the unique characteristics of each attack type. In the text branch, we employ diverse attack text prompts to effectively represent different visual attack images within the aligned text-vision feature space. The following sections will introduce the details of SUEDE.

\subsubsection{Shared Common Feature}
Based on the previous experimental observations and analyses, the text-visual aligned CLIP model is identified as the most suitable foundational network for the Mixture of Experts (MoE) structure. Its robust pre-trained capabilities allow for the effective extraction of common facial structural features and background information across different attack images. 
To leverage this pre-trained visual prior knowledge, the proposed shared expert ($SE$) inherits the parameters from the MLP component of the visual branch in CLIP, which can be expressed as:
\begin{equation}
SE = \text{FFN}_{\text{CLIP}},
\label{eq:shareexpert }
\end{equation}
where $\text{FFN}_{\text{CLIP}}$ denotes the parameters and structure of the FFN layer within the transformer block of the CLIP visual branch. Furthermore, the text prompts utilized by the CLIP model will continue to optimize the shared expert throughout the training process, enabling it to effectively learn common knowledge from various input images. The shared expert is tasked with learning common information across diverse input images, thereby reducing knowledge redundancy and alleviating the optimization burden during model training. In contrast, the routed experts are specifically designed to capture the unique features of various attack images in a personalized manner.

\subsubsection{Diverse Specific Features}
As shown in \cref{fig:algo}, the routed experts in the SUE module are controlled by the gating network to activate and combine their outputs. Specifically, given a feature embedding $x \in \mathbb{R}^{B \times L \times D}$ where $B$ is batch size, $L$ is sequence length, $D$ is hidden dimension, it is input into the gating network $G(x)$ to generates a weight vector $ w = [w_1, w_2, \ldots, w_n] $. Here, $ w_i $ represents the activation weight of the $ i $-th expert, and $\sum_{i=1}^{n} w_i = 1$.
Then, expert $ E_i $ processes the input $x$ to produce the output $ y_i = E_i(x) $. Considering the aforementioned shared expert $SE$, the final output $ y $ of the MoE is computed as the weighted sum of all experts’ outputs, expressed as:
\begin{equation}
y = SE(x) + \sum_{i=1}^{n} w_i \times y_i. \label{eq:sharemoe}
\end{equation}

During the training process, multiple experts can focus on learning the specific features of various attack images, significantly enhancing the model's capability to process combined attack data. The shared expert, augmented by the pre-trained CLIP model, leverages its robust text-visual representation abilities to extract common information across different attack images, thereby reducing redundancy in the learned facial structural knowledge.

\subsubsection{Loss Function}

In addition to CLIP's cross entropy loss, MoE also has a loss function to assist in training \cite{zoph2022stmoe}:

\begin{equation}
    L_{Z}=\,\frac{1}{N}\,\sum_{i=1}^{N}\,\left(\log\sum_{j=1}^{E} \,e^{x^{(i)}_{j}}\right)^{2},
\end{equation}

\noindent where $N$  is the number of tokens, $E$ is the number of experts, and $x \in \mathbb{R}^{B \times N \times E}$ are the logits in the gate network. The loss function $L_Z$ is designed to mitigate the uncertainty of the router, encouraging it to assign each token to a specific expert with greater confidence. By penalizing large normalization constants, it promotes significant differentiation among the gate logits, thereby making the assignments more distinct. Another auxiliary loss $L_B$ is derived from the work of Zoph et al.\cite{zoph2022stmoe}.
Here, $L_B$ means token load balance loss. This loss encourages a balanced distribution of expert utilization, ensuring that no single expert is overutilized. Overall, the final loss is shown as \eqref{loss}. 
In this work, we utilize $\alpha = 1$, $\gamma = 10^{-2}$ and $\beta = 10^{-3}$, which have been empirically shown to effectively aid in training while remaining sufficiently small to avoid overshadowing the primary cross-entropy objective.

\begin{equation}
 L =\alpha  L_{CE} + \beta  L_{Z} + \gamma L_{B}. \label{loss}
\end{equation}

\section{Experiment}

\subsection{Setting}

\noindent{\textbf{Dataset.}}
We employ two large-scale unified physical-digital face attack datasets, including UniAttackData~\cite{fang2024unified} and JFSFDB dataset \cite{yu2024benchmarking}. UniAttackData maintains ID consistency across 2 types of physical attacks and 12 types of digital attacks, involving 1800 subjects. The dataset defines two protocols to comprehensively test the performance of unified attack detection:
1) \textbf{Protocol 1(P1)} aims to evaluate under the unified attack detection task. That is, Protocol 1 includes both physical and digital attacks, with diverse attacks presenting more challenges for algorithm design. 2) \textbf{Protocol 2(P2)} evaluates generalization to ``unseen" attack types, where the significant differences and unpredictability between physical and digital attacks pose challenges to the algorithm's generalization. Protocol 2 is divided into 2 sub-protocols, each with a test set containing an unseen attack type. Protocol 2.1's test set includes only physical attacks unseen in the training and validation sets, while Protocol 2.2's test set includes only digital attacks unseen in the training and validation sets.

JFSFDB dataset \cite{yu2024benchmarking}, compiled by Yu et al., integrating 9 datasets, including SiW\cite{siw}, 3DMAD\cite{3dmad}, HKBU-MarsV2\cite{hkbu}, MSU-MFSD\cite{msu}, 3DMask\cite{3dmask_data}, and ROSE-Youtu\cite{rose} for physical attacks (PAs), as well as FaceForensics++\cite{ff++}, DFDC\cite{dfdc}, and CelebDFv2\cite{cdf} for digital attacks (DAs). This dataset also provides two primary protocols: 1) separate training, where models independently handle the PAs and DAs tasks; 2) joint training, where models simultaneously address the unified attack detection task. In our study, we adopt the joint training scheme, to evaluate the UAD performance.

\noindent{\textbf{Implementation Details.}} The backbone of the image encoder is ViT-B/16 \cite{vit}, text encoder is Transformer\cite{attention}. We set 4 routed experts for each ViT Block, and the number of chosen experts is 2. We explain the reason in the ablation study later. The training process is running on an NVIDIA RTX3090 GPU using the Adam optimizer, with an initial learning rate of $10^{-6}$.

\subsection{Performance}

\setlength{\tabcolsep}{10mm}{
\begin{table}[ht]
\vspace{-4mm}
\renewcommand\arraystretch{1} 
\centering
\setlength{\tabcolsep}{1mm}
\resizebox{1\linewidth}{!}{
\begin{tabular}{c|c|cccc}

\toprule[1pt] 
\textbf{Prot.}     & \textbf{Method}         & \textbf{ACER(\%)$\downarrow$}    & \textbf{ACC(\%)$\uparrow$}     & \textbf{AUC(\%)$\uparrow$}    & \textbf{EER(\%)$\downarrow$}    \\ \midrule[1pt]
\multirow{6}{*}{\textbf{1}} 
                   & ResNet50                & 1.35                 & 98.83                & 99.79               & 1.18                \\
                   & VIT-B/16                & 5.92                 & 92.29                & 97.00               & 9.14                \\
                   & Auxiliary               & 1.13                 & 98.68                & 99.82               & 1.23              \\
                   & CDCN                    & 1.40                 & 98.57                & 99.52               & 1.42                \\
                   & FFD                     & 2.01                 & 97.97                & 99.57               & 2.01                \\
                  & CLIP         & 1.17              & 99.13            & 99.66            & 1.17             \\
                   & UniAttackDetection      & 0.52                & 99.45                & 99.96               & 0.53                \\
                   \rowcolor{mygray}
                   & \textbf{SUEDE} & \textbf{0.36}        & \textbf{99.50}       & \textbf{99.70}      & \textbf{0.36}       \\ \midrule[1pt]
\multirow{6}{*}{\textbf{2}} & ResNet50                & 34.60$\pm$5.31           & 53.69$\pm$6.39           & 87.89$\pm$6.11          & 19.48$\pm$9.10         \\
                   & VIT-B/16                & 33.69$\pm$9.33           & 52.43$\pm$25.88          & 83.77$\pm$2.35          & 25.94$\pm$ 0.88          \\
                   & Auxiliary               & 42.98$\pm$6.77           & 37.71$\pm$26.45          & 76.27$\pm$12.06         & 32.66$\pm$7.91          \\
                   & CDCN                    & 34.33$\pm$0.66           & 53.10$\pm$12.70          & 77.46$\pm$17.56         & 29.17$\pm$14.47         \\
                   & FFD                     & 44.20$\pm$1.32            & 40.43$\pm$14.88         & 80.97$\pm$2.86          & 26.18$\pm$2.77               \\
                   & CLIP                     & 18.20$\pm$13.72            & 54.94$\pm$20.38         & 84.09$\pm$14.96          & 18.05$\pm$13.86              \\
                   & UniAttackDetection                   &22.42$\pm$10.57          & 67.35$\pm$23.22         & 91.97$\pm$4.55          & 15.72$\pm$3.08               \\
                   \rowcolor{mygray}
                   & \textbf{SUEDE} & \textbf{11.99$\pm$7.79} & \textbf{88.99$\pm$6.96} & \textbf{91.49$\pm$4.99} & \textbf{11.89$\pm$7.88} \\ \bottomrule[1pt]
\end{tabular}
}
\caption{The results on two protocols of UniAttackData, where the performance of Protocol 2 quantified as the mean$\pm$std measure derived from Protocol 2.1 and Protocol 2.2.}
\label{Table:p1}
\vspace{-2mm}
\end{table}
}

To evaluate the performance of our proposed method, we selected the Average Classification Error Rate (ACER), Accuracy (ACC), Area Under the Curve (AUC), and Equal Error Rate (EER) as performance evaluation metrics. We compared our method with ResNet50 \cite{resnet}, ViT-B/16 \cite{vit}, FFD \cite{ffd}, CLIP \cite{radford2021CLIP}, CDCN \cite{yu2020cdcn}, Auxiliary \cite{siw}, and UniAttackDetection \cite{fang2024unified}. \cref{Table:p1} shows the results of UniAttackData \cite{fang2024unified}, and our proposed method has the best ACER and ACC, which means it effectively detects fraudulent behavior. Next, we will provide a detailed explanation of the experimental results of the two protocols.

\noindent\textbf{Expertiment on protocol 1.}
In Protocol 1 of UniAttackData, the training, validation, and test sets all contain both physical and digital attacks. This protocol is used to evaluate the performance of algorithms in the unified attack detection task. We present the performance results of commonly used backbone networks, physical attack detection networks, and digital attack detection networks. As shown in \cref{Table:p1}, 
Our method achieves the best result: ACER is 0.36\%, ACC is 99.50\%, AUC is 99.70\%, and EER is 0.36\%. 
Next, we visualize the feature distribution using SUEDE in in \cref{fig:tsne}(a), which shows that SUEDE can learn unified features across both physical attacks and digital attacks for identifying fake faces. Both the experiments and visualizations indicate that SUEDE effectively detects the both attack types simultaneously, demonstrating the superiority of our proposed shared unified experts.

\noindent\textbf{Expertiment on protocol 2.}
In the Protocol 2 of the UniAttackData dataset, the test set includes ``unseen" attack types that were not present in the training or validation sets, aiming to evaluate the generalization capabilities to unknown domains. As shown in \cref{Table:p1}, the state-of-the-art (SOTA) unified attack detection (UniAttackDetection) and single attack detection methods (FFD, CDCN, Auxiliary) all experience a significant drop in performance under the unseen setting. In contrast, our method still achieves the best performance across all metrics, with a significant margin over the other methods, demonstrating its superior generalization capabilities.

\setlength{\tabcolsep}{10mm}{
\begin{table}[ht]
\renewcommand\arraystretch{1} 
\centering
\setlength{\tabcolsep}{1mm}
\resizebox{1.0\linewidth}{!}{
\begin{tabular}{c|c|cccc}
      \toprule[1pt] 
      \textbf{Data.} & 
      \textbf{Method} & 
      \textbf{AUC(\%)$\uparrow$} & \textbf{EER(\%)$\downarrow$} &
      \textbf{TPR(\%)@FPR=10\%$\uparrow$} & \textbf{TPR(\%)@FPR=1\%$\uparrow$} \\
      \midrule[1pt]
      & ResNet50 & 82.38 & 24.62 & 57.66 & 21.72 \\
      & VIT-B/16 & 82.70 & 24.60 & 61.67 & 22.60 \\
      & DeepPixel & 78.00 & 28.67 & 49.98 & 18.60 \\
      & CDCN & 70.04 & 36.64 & 45.43 & 17.46 \\
      & MultiAtten & 69.36 & 35.21 & 36.88 & 11.37 \\
      & Xception & 77.80 & 27.62 & 44.58 & 16.99 \\
      & MesoNet & 61.09 & 42.11 & 24.85 & 5.77 \\
      \multirow{-6}{*}{JFSFDB} & \cellcolor{mygray}\textbf{SUEDE} & \cellcolor{mygray}\textbf{87.02} & \cellcolor{mygray}\textbf{21.42} & \cellcolor{mygray}\textbf{62.29} & \cellcolor{mygray}\textbf{24.70} \\
      \bottomrule[1pt] 
    \end{tabular}
}
\caption{The results on joint training(unified attack data) protocol of JFSFDB.}
\label{Table:jfs_cross}
\vspace{-4mm}
\end{table}
}

To further evaluate the effectiveness of the proposed method, we conduct experiments on a larger dataset JFSFDB \cite{yu2024benchmarking} using cross domain testing in the joint training protocol. we also add several methods for comparison: DeepPixel \cite{george2019DeepPixel}, MultiAtten \cite{zhao2021multiAtten}, Xception \cite{ff++}, and MesoNet \cite{afchar2018mesonet}.
As shown in \cref{Table:jfs_cross}, SUEDE surpasses other baselines by a big margin.

\subsection{Ablation Experiment}

\noindent \textbf{Different expert settings.} We first evaluate the effectiveness of embedding the SUE module across different modalities(visual/text). We conduct experiment on Protocol 1 of the UniAttackData dataset. As reported in~\cref{Table:ablation-1}, employing the vanilla MoE, which lacks a shared expert, results in significant performance degradation. In contrast, embedding the SUE module across different modalities leads to substantial performance improvements, highlighting the superiority of the SUE module. Notably, embedding SUE in the visual encoder achieves the best performance. Therefore, we adopt this configuration as the default setting.

\begin{table}[ht]
\centering
\renewcommand\arraystretch{1} 
\setlength{\tabcolsep}{1mm} 
\resizebox{0.9\linewidth}{!}{
\begin{tabular}{c|cccc}
\toprule[1pt] 
\textbf{Method}           & \textbf{ACER(\%)}$\downarrow$ & \textbf{ACC(\%)}$\uparrow$ & \textbf{AUC(\%)$\uparrow$} & \textbf{EER(\%)}$\downarrow$ \\ \midrule[1pt]
visual MoE        & 13.03            & 92.56            & 97.21            & 9.1             \\
text SUE        & 0.73              & 99.32            & 99.67            & 0.73             \\
text\&visual SUE       &0.87              & 99.14           & 99.57            & 0.87             \\

\rowcolor{mygray}
\textbf{visual SUE}   & \textbf{0.36}      & \textbf{99.50}   & \textbf{99.70}   & \textbf{0.36}    \\ \bottomrule[1pt]
\end{tabular}
}
\caption{The ablation study of different expert settings. The evaluation protocol is Protocol 1. SUE means our proposed Shared Unified Experts. }
\label{Table:ablation-1}
\vspace{-2mm}
\end{table}

\begin{figure}[t]
\centering
\includegraphics[width=1.0\linewidth]{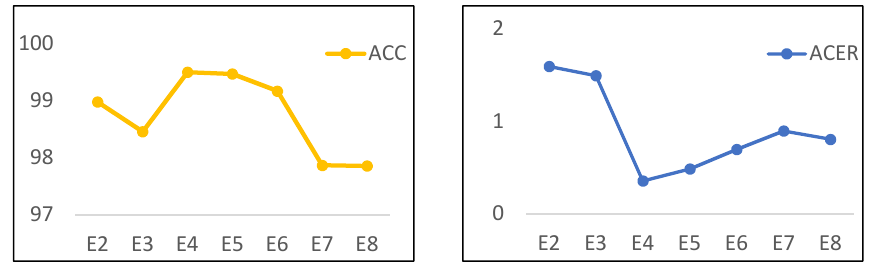}
\caption{The ablation experiment with the number of experts.The horizontal axis represents the number of experts from 2 to 8, while the vertical axis represents ACC (\%) and ACER (\%).}
\label{fig:numOfE}
\vspace{-6mm}
\end{figure}

\noindent \textbf{The number of experts.} We also examine the effect of the number of routed experts on performance. The number of routed experts is increased incrementally from 2 to 8, selecting 2 experts at each step. As shown in \cref{fig:numOfE}, the best performance, measured by metrics such as ACER and ACC, is achieved when the number of routed experts is set to 4.

\begin{table}[!ht]
\centering
\renewcommand\arraystretch{1} 
\setlength{\tabcolsep}{1mm} 
\resizebox{0.7\linewidth}{!}{
\begin{tabular}{c|cccc}
\toprule[1pt] 
\textbf{Method}           & \textbf{ACER(\%)}$\downarrow$ & \textbf{ACC(\%)}$\uparrow$ & \textbf{AUC(\%)$\uparrow$} & \textbf{EER(\%)}$\downarrow$ \\ \midrule[1pt]

$MoE_{0-2}$         & 0.77              & 99.11            & 99.51            & 0.77             \\
$MoE_{5-7}$         & 0.76              & 99.37            & 99.72            & 0.66             \\
$MoE_{9-11}$        & 0.42              & 99.53           & 99.75            & 0.42            \\
\rowcolor{mygray}
\textbf{$MoE_{0-11}$}   & \textbf{0.36}      & \textbf{99.50}   & \textbf{99.70}   & \textbf{0.36}    \\ \bottomrule[1pt]
\end{tabular}
}
\caption{The benefit of experts embedding in different layers. The evaluation protocol is P1.}
\label{Table:ablation-2}
\vspace{-2mm}
\end{table}

\noindent \textbf{The benefit of experts embedding in different layers.} In this ablation study, we evaluate the detection performance of placing experts at different layers within the visual encoder. In \cref{Table:ablation-2}, $MoE_{0-2}$  signifies the incorporation of MoE in layers (0, 1, 2), representing the effect of MoE in the early stages of the encoder. The remaining three configurations denote the integration of MoE in the middle, later stages, and throughout the entire encoder (the Shared MoE setting). 
It is evident that incorporating MoE in every layer and integrating MoE in the later stages of the encoder have superior detection performance.

\section{Conclusion}

In this work, we propose shared unified experts for physical-
digital face attack detection enhancement. It solves the previous problem of deploying multiple models to deal with different attacks and optimizes the detection of diverse and complex unified attacks. We conducted experiments on the UniAttackData and JFSFDB datasets, and detailed experimental results demonstrated the effectiveness of our method for unified attack detection(UAD) tasks.





\section*{Acknowledgment}
This work is supported by the National Natural Science Foundations of China (62302139, 62406320, 62272144), the Fundamental Research Funds for the Central Universities (JZ2023HGTA0202, JZ2023HGQA0101, JZ2024HGTG0309, JZ2024AHST0337), the National Key R\&D Program of China (NO.2024YFB3311602), the Anhui Provincial Natural Science Foundation (2408085J040), and the Major Project of Anhui Provincial Science and Technology Breakthrough Program (202423k09020001). 

\bibliographystyle{IEEEbib}
\bibliography{main}

\begin{thebibliography}{10}

\bibitem{liu2021casia}
Ajian Liu, Zichang Tan, Jun Wan, Sergio Escalera, Guodong Guo, and Stan~Z Li,
\newblock ``Casia-surf cefa: A benchmark for multi-modal cross-ethnicity face anti-spoofing,''
\newblock in {\em WACV}, 2021, pp. 1179--1187.

\bibitem{liu2022contrastive}
Ajian Liu, Chenxu Zhao, Zitong Yu, Jun Wan, Anyang Su, Xing Liu, Zichang Tan, Sergio Escalera, Junliang Xing, Yanyan Liang, et~al.,
\newblock ``Contrastive context-aware learning for 3d high-fidelity mask face presentation attack detection,''
\newblock {\em TIFS}, vol. 17, pp. 2497--2507, 2022.

\bibitem{cdf}
Yuezun Li, Xin Yang, Pu~Sun, Honggang Qi, and Siwei Lyu,
\newblock ``Celeb-df: A large-scale challenging dataset for deepfake forensics,''
\newblock in {\em CVPR}, 2020, pp. 3207--3216.

\bibitem{2d}
Raghavendra Ramachandra and Christoph Busch,
\newblock ``Presentation attack detection methods for face recognition systems: A comprehensive survey,''
\newblock {\em ACM Computing Surveys (CSUR)}, vol. 50, no. 1, pp. 1--37, 2017.

\bibitem{ijcai2022p165}
Ajian Liu and Yanyan Liang,
\newblock ``Ma-vit: Modality-agnostic vision transformers for face anti-spoofing,''
\newblock in {\em IJCAI-22}, 2022, pp. 1180--1186.

\bibitem{Liu2023FMViTFM}
Ajian Liu, Zichang Tan, Zitong Yu, Chenxu Zhao, Jun Wan, Yanyan Liang, Zhen Lei, Du~Zhang, S.~Li, and Guodong Guo,
\newblock ``Fm-vit: Flexible modal vision transformers for face anti-spoofing,''
\newblock {\em IEEE Transactions on Information Forensics and Security}, vol. 18, pp. 4775--4786, 2023.

\bibitem{liu2021face}
Ajian Liu, Zichang Tan, Jun Wan, Yanyan Liang, Zhen Lei, Guodong Guo, and Stan~Z Li,
\newblock ``Face anti-spoofing via adversarial cross-modality translation,''
\newblock {\em TIFS}, vol. 16, pp. 2759--2772, 2021.

\bibitem{ciftci2021detection}
Umur~Aybars Ciftci, Ilke Demir, and Lijun Yin,
\newblock ``Detection of synthetic portrait videos using biological signals,''
\newblock {\em arXiv}, 2021.

\bibitem{simswap}
Renwang Chen, Xuanhong Chen, Bingbing Ni, and Yanhao Ge,
\newblock ``Simswap: An efficient framework for high fidelity face swapping,''
\newblock in {\em {MM} '20: The 28th {ACM} International Conference on Multimedia}, 2020.

\bibitem{fd_forgeryArea}
Huy~H Nguyen, Fuming Fang, Junichi Yamagishi, and Isao Echizen,
\newblock ``Multi-task learning for detecting and segmenting manipulated facial images and videos,''
\newblock in {\em BTAS}. IEEE, 2019, pp. 1--8.

\bibitem{mct_location}
Changtao Miao, Qi~Chu, Weihai Li, Suichan Li, Zhentao Tan, Wanyi Zhuang, and Nenghai Yu,
\newblock ``Learning forgery region-aware and id-independent features for face manipulation detection,''
\newblock {\em TBIOM}, vol. 4, no. 1, pp. 71--84, 2022.

\bibitem{fd_disentagled}
Jiahao Liang, Huafeng Shi, and Weihong Deng,
\newblock ``Exploring disentangled content information for face forgery detection,''
\newblock in {\em ECCV}. Springer, 2022, pp. 128--145.

\bibitem{fd_imgRecon}
Junyi Cao, Chao Ma, Taiping Yao, Shen Chen, Shouhong Ding, and Xiaokang Yang,
\newblock ``End-to-end reconstruction-classification learning for face forgery detection,''
\newblock in {\em CVPR}, 2022, pp. 4113--4122.

\bibitem{fd_f1}
Ricard Durall, Margret Keuper, and Janis Keuper,
\newblock ``Watch your up-convolution: Cnn based generative deep neural networks are failing to reproduce spectral distributions,''
\newblock in {\em CVPR}, 2020, pp. 7890--7899.

\bibitem{mct_frequency_1}
Changtao Miao, Zichang Tan, Qi~Chu, Huan Liu, Honggang Hu, and Nenghai Yu,
\newblock ``F2trans: High-frequency fine-grained transformer for face forgery detection,''
\newblock {\em TIFS}, vol. 18, pp. 1039--1051, 2023.

\bibitem{mct_frequency_2}
Changtao Miao, Zichang Tan, Qi~Chu, Nenghai Yu, and Guodong Guo,
\newblock ``Hierarchical frequency-assisted interactive networks for face manipulation detection,''
\newblock {\em IEEE Transactions on Information Forensics and Security}, vol. 17, pp. 3008--3021, 2022.

\bibitem{liu2024moeit}
Ajian Liu,
\newblock ``Ca-moeit: Generalizable face anti-spoofing via dual cross-attention and semi-fixed mixture-of-expert,''
\newblock {\em International Journal of Computer Vision}, pp. 1--14, 2024.

\bibitem{ffd}
Hao Dang, Feng Liu, Joel Stehouwer, Xiaoming Liu, and Anil~K Jain,
\newblock ``On the detection of digital face manipulation,''
\newblock in {\em CVPR}, 2020, pp. 5781--5790.

\bibitem{fang2024unified}
Hao Fang, Ajian Liu, Haocheng Yuan, Junze Zheng, Dingheng Zeng, Yanhong Liu, Jiankang Deng, Sergio Escalera, Xiaoming Liu, Jun Wan, et~al.,
\newblock ``Unified physical-digital face attack detection,''
\newblock {\em arXiv preprint arXiv:2401.17699}, 2024.

\bibitem{liu2024cfpl}
Ajian Liu, Shuai Xue, Jianwen Gan, Jun Wan, Yanyan Liang, Jiankang Deng, Sergio Escalera, and Zhen Lei,
\newblock ``Cfpl-fas: Class free prompt learning for generalizable face anti-spoofing,''
\newblock in {\em CVPR}, 2024, pp. 222--232.

\bibitem{yu2024benchmarking}
Zitong Yu, Rizhao Cai, Zhi Li, Wenhan Yang, Jingang Shi, and Alex~C Kot,
\newblock ``Benchmarking joint face spoofing and forgery detection with visual and physiological cues,''
\newblock {\em IEEE Transactions on Dependable and Secure Computing}, 2024.

\bibitem{zou2024softmoe}
Hang Zou, Chenxi Du, Hui Zhang, Yuan Zhang, Ajian Liu, Jun Wan, and Zhen Lei,
\newblock ``La-softmoe clip for unified physical-digital face attack detection,''
\newblock in {\em 2024 IEEE International Joint Conference on Biometrics (IJCB)}. IEEE, 2024, pp. 1--11.

\bibitem{jacobs1991moe1}
Robert~A Jacobs, Michael~I Jordan, Steven~J Nowlan, and Geoffrey~E Hinton,
\newblock ``Adaptive mixtures of local experts,''
\newblock {\em Neural computation}, vol. 3, no. 1, pp. 79--87, 1991.

\bibitem{dai2024deepseekmoe}
Damai Dai, Chengqi Deng, Chenggang Zhao, R.~X. Xu, Huazuo Gao, Deli Chen, Jiashi Li, Wangding Zeng, Xingkai Yu, Y.~Wu, Zhenda Xie, Y.~K. Li, Panpan Huang, Fuli Luo, Chong Ruan, Zhifang Sui, and Wenfeng Liang,
\newblock ``Deepseekmoe: Towards ultimate expert specialization in mixture-of-experts language models,''
\newblock {\em CoRR}, vol. abs/2401.06066, 2024.

\bibitem{shazeer2017outrageously}
Noam Shazeer, Azalia Mirhoseini, Krzysztof Maziarz, Andy Davis, Quoc Le, Geoffrey Hinton, and Jeff Dean,
\newblock ``Outrageously large neural networks: The sparsely-gated mixture-of-experts layer,''
\newblock {\em arXiv preprint arXiv:1701.06538}, 2017.

\bibitem{radford2021CLIP}
Alec Radford, Jong~Wook Kim, Chris Hallacy, Aditya Ramesh, Gabriel Goh, Sandhini Agarwal, Girish Sastry, Amanda Askell, Pamela Mishkin, Jack Clark, et~al.,
\newblock ``Learning transferable visual models from natural language supervision,''
\newblock in {\em International conference on machine learning}. PMLR, 2021, pp. 8748--8763.

\bibitem{tsne}
Laurens Van~der Maaten and Geoffrey Hinton,
\newblock ``Visualizing data using t-sne.,''
\newblock {\em Journal of machine learning research}, vol. 9, no. 11, 2008.

\bibitem{zoph2022stmoe}
Barret Zoph, Irwan Bello, Sameer Kumar, Nan Du, Yanping Huang, Jeff Dean, Noam Shazeer, and William Fedus,
\newblock ``St-moe: Designing stable and transferable sparse expert models,''
\newblock {\em arXiv preprint arXiv:2202.08906}, 2022.

\bibitem{siw}
Yaojie Liu, Amin Jourabloo, and Xiaoming Liu,
\newblock ``Learning deep models for face anti-spoofing: Binary or auxiliary supervision,''
\newblock in {\em Proceedings of the IEEE CVPR}, 2018, pp. 389--398.

\bibitem{3dmad}
Nesli Erdogmus and Sebastien Marcel,
\newblock ``Spoofing face recognition with 3d masks,''
\newblock {\em IEEE transactions on information forensics and security}, vol. 9, no. 7, pp. 1084--1097, 2014.

\bibitem{hkbu}
Siqi Liu, Pong~C Yuen, Shengping Zhang, and Guoying Zhao,
\newblock ``3d mask face anti-spoofing with remote photoplethysmography,''
\newblock in {\em Computer Vision--ECCV 2016: 14th European Conference, Amsterdam, The Netherlands, October 11--14, 2016, Proceedings, Part VII 14}. Springer, 2016, pp. 85--100.

\bibitem{msu}
Di~Wen, Hu~Han, and Anil~K Jain,
\newblock ``Face spoof detection with image distortion analysis,''
\newblock {\em IEEE Transactions on Information Forensics and Security}, vol. 10, no. 4, pp. 746--761, 2015.

\bibitem{3dmask_data}
Zitong Yu, Jun Wan, Yunxiao Qin, Xiaobai Li, Stan~Z Li, and Guoying Zhao,
\newblock ``Nas-fas: Static-dynamic central difference network search for face anti-spoofing,''
\newblock {\em IEEE transactions on pattern analysis and machine intelligence}, vol. 43, no. 9, pp. 3005--3023, 2020.

\bibitem{rose}
Haoliang Li, Wen Li, Hong Cao, Shiqi Wang, Feiyue Huang, and Alex~C Kot,
\newblock ``Unsupervised domain adaptation for face anti-spoofing,''
\newblock {\em IEEE Transactions on Information Forensics and Security}, vol. 13, no. 7, pp. 1794--1809, 2018.

\bibitem{ff++}
Andreas Rossler, Davide Cozzolino, Luisa Verdoliva, Christian Riess, Justus Thies, and Matthias Nie{\ss}ner,
\newblock ``Faceforensics++: Learning to detect manipulated facial images,''
\newblock in {\em ICCV}, 2019, pp. 1--11.

\bibitem{dfdc}
B~Dolhansky,
\newblock ``The dee pfake detection challenge (dfdc) pre view dataset,''
\newblock {\em arXiv preprint arXiv:1910.08854}, 2019.

\bibitem{vit}
Alexey Dosovitskiy,
\newblock ``An image is worth 16x16 words: Transformers for image recognition at scale,''
\newblock {\em arXiv preprint arXiv:2010.11929}, 2020.

\bibitem{attention}
A~Vaswani,
\newblock ``Attention is all you need,''
\newblock {\em Advances in Neural Information Processing Systems}, 2017.

\bibitem{resnet}
Kaiming He, Xiangyu Zhang, Shaoqing Ren, and Jian Sun,
\newblock ``Deep residual learning for image recognition,''
\newblock in {\em CVPR}, 2016, pp. 770--778.

\bibitem{yu2020cdcn}
Zitong Yu, Chenxu Zhao, Zezheng Wang, Yunxiao Qin, Zhuo Su, Xiaobai Li, Feng Zhou, and Guoying Zhao,
\newblock ``Searching central difference convolutional networks for face anti-spoofing,''
\newblock in {\em CVPR}, 2020, pp. 5295--5305.

\bibitem{george2019DeepPixel}
Anjith George and S{\'e}bastien Marcel,
\newblock ``Deep pixel-wise binary supervision for face presentation attack detection,''
\newblock in {\em 2019 International Conference on Biometrics (ICB)}. IEEE, 2019, pp. 1--8.

\bibitem{zhao2021multiAtten}
Hanqing Zhao, Wenbo Zhou, Dongdong Chen, Tianyi Wei, Weiming Zhang, and Nenghai Yu,
\newblock ``Multi-attentional deepfake detection,''
\newblock in {\em CVPR}, 2021, pp. 2185--2194.

\bibitem{afchar2018mesonet}
Darius Afchar, Vincent Nozick, Junichi Yamagishi, and Isao Echizen,
\newblock ``Mesonet: a compact facial video forgery detection network,''
\newblock in {\em 2018 IEEE international workshop on information forensics and security (WIFS)}. IEEE, 2018, pp. 1--7.

\end{thebibliography}

\end{document}